\documentclass[10pt,journal,compsoc]{IEEEtran}

\usepackage{color}

\ifCLASSOPTIONcompsoc
\usepackage[nocompress]{cite}
\else
\usepackage{cite}
\fi

\usepackage{ulem}
\usepackage{times}
\usepackage{epsfig}
\usepackage{graphicx}
\usepackage{amsmath}
\usepackage{amssymb}
\usepackage{amsmath}

\usepackage{algorithm}
\usepackage{algorithmic}

\usepackage{graphicx} 
\graphicspath{{figs/}}

\begin{document}
%
\title{Unsupervised Partial Point Set Registration via Joint Shape Completion and Registration}

\author{Xiang~Li,
Lingjing Wang,
and~Yi~Fang$\dagger$
\thanks{Xiang Li, Lingjing Wang, and Yi Fang are with NYU Multimedia and Visual Computing Lab, NYU Tandon and Abu Dhabi, also with the Department of Electrical and Computer Engineering, NYU Tandon, USA, and with the Department of Electrical and Computer Engineering, NYU Abu Dhabi, UAE. Email:\{xl1845,lw1474,yfang\}@nyu.edu.}
\thanks{$\dagger$ Corresponding author: Yi Fang (yfang@nyu.edu).}
}

\markboth{IEEE TRANSACTIONS ON VISUALIZATION AND COMPUTER GRAPHICS}{}

\IEEEtitleabstractindextext{%
\begin{abstract}
We propose a self-supervised method for partial point set registration. While recent proposed learning-based methods have achieved impressive registration performance on the full shape observations, these methods mostly suffer from performance degradation when dealing with partial shapes. To bridge the performance gaps between partial point set registration with full point set registration, we proposed to incorporate a shape completion network to benefit the registration process. To achieve this, we design a latent code for each pair of shapes, which can be regarded as a geometric encoding of the target shape. By doing so, our model does need an explicit feature embedding network to learn the feature encodings. More importantly, both our shape completion network and the point set registration network take the shared latent codes as input, which are optimized along with the parameters of two decoder networks in the training process. Therefore, the point set registration process can thus benefit from the joint optimization process of latent codes, which are enforced to represent the information of full shape instead of partial ones. In the inference stage, we fix the network parameter and optimize the latent codes to get the optimal shape completion and registration results. Our proposed method is pure unsupervised and does not need any ground truth supervision. Experiments on the ModelNet40 dataset demonstrate the effectiveness of our model for partial point set registration. 
\end{abstract}

\begin{IEEEkeywords}
Point Set Registration, Partial Registration, Unsupervised learning, Shape Completion.
\end{IEEEkeywords}}

\maketitle

\IEEEdisplaynontitleabstractindextext

%
\IEEEpeerreviewmaketitle

\IEEEraisesectionheading{\section{Introduction}\label{sec:introduction}}
\IEEEPARstart{P}{oint} set registration is generally defined as estimating the spatial transformation—either rigid or non-rigid transformation to align one point set to another. It is a key component in various applications such as robotics, shape correspondence, and large-scale 3D reconstruction \cite{wang2019deep}. During the past decades, the point set registration problem has been an active research area in the computer vision community, with a large number of both traditional and recent approaches \cite{besl1992method,yang2015go,zhou2016fast,wang2019deep,wang2019prnet}.

Given a pair of point sets, traditional methods mostly solve the registration problem by iteratively optimizing a pre-defined alignment loss between the transformed source point sets and their corresponding target ones. These methods, however, need to start over a new optimization process for each new pair of inputs and require an individual tuning process for the hyperparameters for each case. Therefore, the trade-off between efficiency and effectiveness becomes one major concern for traditional methods.

In recent years, with the prevalence of deep learning in various vision tasks, researchers have been shifting their focus from ``case-driven'' iterative-based methods towards ``data-driven" learning-based methods for point set registration. Recently proposed methods, such as PointNetLK \cite{aoki2019pointnetlk} and Deep Closest Point (DCP) \cite{wang2019deep}, show that deep neural networks with strong non-linear modeling capacity have superior performance in comparison to classical methods. After training, deep learning-based methods can produce high-quality registration outputs in one network forward pass.  Most importantly, these methods show good generalization abilities when transferred to unseen pairs, even when trained on different datasets. Learning-based methods greatly enhance the registration efficiency and enable the possibility of registration of large datasets. Despite the great successes achieved, these methods suffer from performance degradation when the input point sets have missing parts.

\begin{figure*}[h]
    \centering
    \includegraphics[width=17cm]{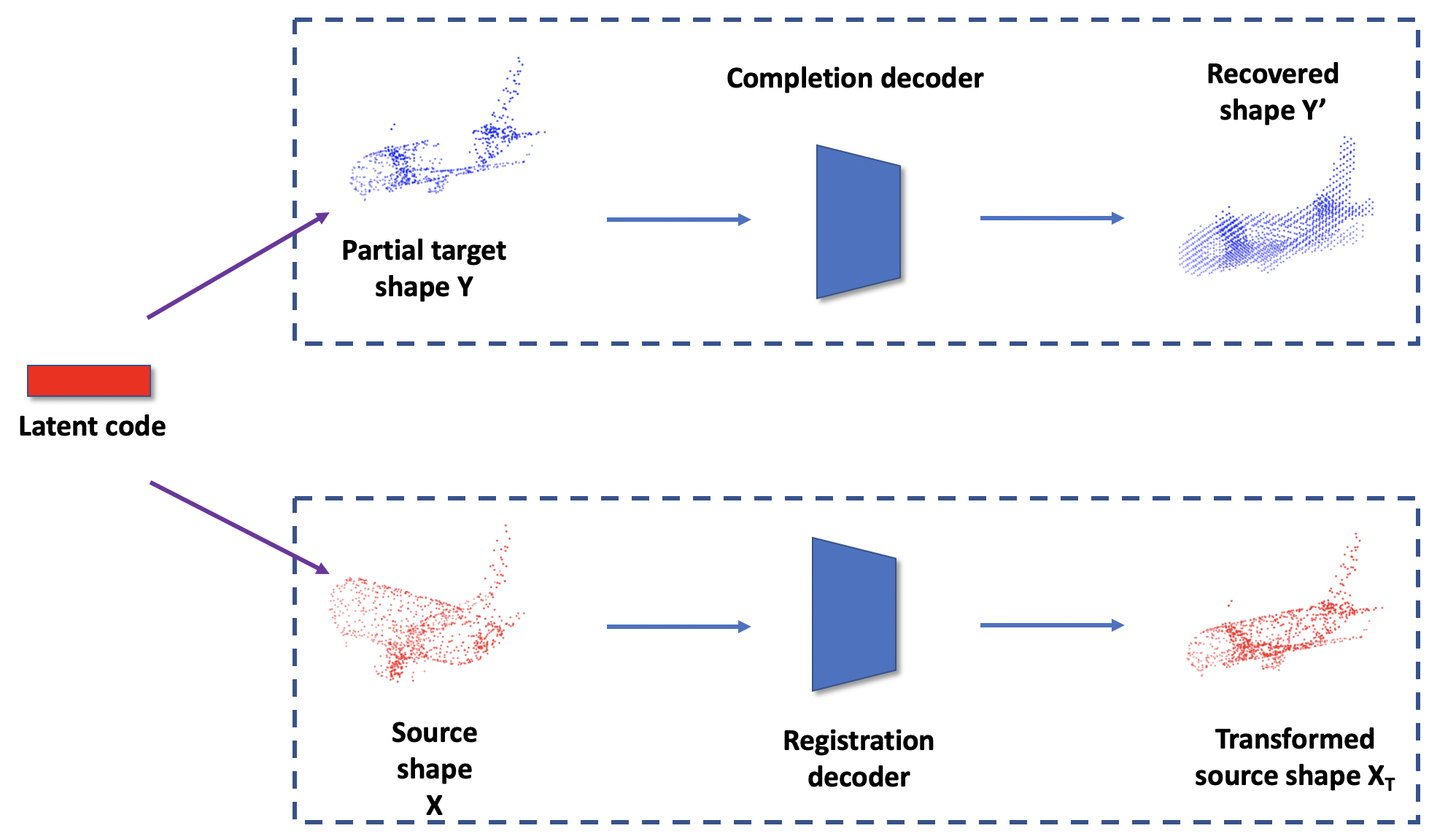}
    \caption{Our proposed method for partial point set registration. Our model achieves simultaneously shape completion and point set registration.}
    \label{fig_intro}
\end{figure*}

More recently, in \cite{balakrishnan2018unsupervised}, the author proposes probably the first method for partial point set registration. The authors firstly detect keypoints from partial observations and then learns a mapping from keypoints of source shape to keypoints of the target shape as the desired transformation. This method needs to refine the alignment in an iterative process, which makes the method time-consuming. Moreover, their registration performance highly depends on the performance of keypoint detection.

In this paper, we approach the partially point set registration task from a different perspective. Our method is motivated by two general findings: 1) the problem of point set registration on full point sets is a well-studied problem with satisfying and significantly better performance than the problem of partial point set registration. For example, the recently proposed DCP \cite{wang2019deep} method reports a mean square error of 45.01 and 1.31 for rotation angle prediction on the ModelNet40 partial and full point sets, respectively; 2) recent progress in deep learning fields has seen a lot of learning-based methods for shape completion from partial observations. These findings naturally motivate us to handle the partial point set registration problem to recover full shapes from partial inputs while performing point set registration. Therefore, in this paper, we proposed to achieve partial point set registration by jointly optimizing two tasks: 1) to recover full 3D shapes from the input partial point sets, and 2) to learn a rigid transformation from source point set to the target point set. See Figure \ref{fig_intro} for illustration.

In order to integrate these two tasks for joint optimization, we proposed to incorporate a latent code for each pair of shapes, which captures the geometric essence of the input shapes. Both our shape completion and point set registration network take a shared latent code as input and optimizes it during network training. Our shape completion network enforces the optimized latent codes to represent the information of full shapes instead of partial inputs. Therefore, our point set registration network can benefit from the shape completion network during the joint optimization of the shared latent code. Our contributions are listed as below:

\begin{itemize}
\item We propose a novel unsupervised point set registration framework that enables partially point cloud registration using deep neural networks.
\item  To eliminate the explicit design of the feature encoding network, we formulate a learnable latent vector for each shape as the global feature representation.
\item Our model achieves simultaneously shape completion and point set registration by optimizing a shared latent vector. The point set registration network can thus benefit from the shape completion network that enforces the learned latent code to represent the information of full shape instead of partial ones.
\item Experiments on the ModelNet40 benchmark dataset demonstrate the effectiveness of our model for partial point set registration. 
\end{itemize}

\section{Related Works}
\subsection{Point Set Registration}
Point sets registration is generally defined as finding the geometric transformations to align the source point set with the target one. Classical iterative optimization methods usually search a set of optimal parameters of a transformation during the process of minimization of a pre-defined alignment loss. The Iterative Closest Point (ICP) algorithm \cite{besl1992method} is one of the classical solutions which achieves great success for rigid point set registration. ICP leverages a set of corresponding points to define the transformation and iteratively refine the transformation by minimizing an error metric. In ICP, the initial guess of the desired rigid transformation can have a large impact on the final performance. Yang et al. \cite{yang2015go} propose Go-ICP to release the local initialization problem of ICP by using a BnB searching scheme over the entire 3D motion space. Other methods such as TPS-RSM \cite{chui2000new} and CPD \cite{myronenko2007non} propose solutions to solve non-rigid registration problem. One common disadvantage of these transitional methods is about registration efficiency. All these methods register every single pair of source and target point sets as an independent optimization process, which cannot transfer knowledge from registering one pair to another. 

Recently, deep learning methods have achieved great success in many computer vision applications, including image classification \cite{simonyan2014very,he2016deep}, semantic segmentation \cite{long2015fully,li2018building}, object detection \cite{ren2015faster,redmon2016you,hu2019sample}, image registration \cite{rocco2017convolutional,chen2019arbicon}, point signature learning \cite{qi2017pointnet,masci2015geodesic,li2020dance,wang2020few}, etc. Many researchers shift their attention from classical methods to learning-based methods to approach the point set registration problem \cite{rocco2017convolutional,balakrishnan2018unsupervised,zeng20173dmatch,qi2017pointnet,verma2018feastnet,masci2015geodesic}. Aoki et al. \cite{aoki2019pointnetlk} propose PointNetLK to combine the Lucas $\&$ Kanade algorithm with the PointNet feature exaction module into a single trainable recurrent deep neural network. Liu et al. propose FlowNet3D \cite{liu2019flownet3d}, which predicts the flow field to move the source point set towards the target one. Deep Closest Point \cite{wang2019deep} leverages a DGCNN structure for point feature learning, followed by an attention-based feature matching module to generate the correspondence. Wang et al. \cite{wang2019prnet} proposed PR-Net for learning the partially rigid registration. PR-Net firstly leverages a keypoint detector to find the corresponding points of the source and target point sets. The keypoint is determined based on the per-point feature embeddings of input shapes. Then based on the matching keypoints, the desired mapping can be further predicted. Unlike DCP and PR-Net that both use an explicit point feature encoding network to learn per-point features, in our model, we represent the geometric features of 3D shapes using an optimizable latent vector and thus enable joint shape completion and point set registration.


\subsection{Shape Completion}
3D shape completion is generally defined as recovering the unseen parts of the original partial observations. It's a long-standing problem in computer vision and graphics fields with extensive researches, including both traditional non-learning based methods and recent learning-based methods. Traditional shape completion methods fill the unseen parts by assuming local surface or volumetric smoothness. For example, one can fill holes with surface primitives, such as planes or quadrics, or casts the problem as an energy minimization, e.g., Laplacian smoothing \cite{sorkine2004least,zhao2007robust}. Other researches leverage observed structures and regularities in 3D shapes, such as symmetries, to complete shapes \cite{thrun2005shape,pauly2008discovering}.  Another commonly used strategy is to search a most similar template shape from a database and align it with partial observations to achieve shape completion \cite{kraevoy2005template,Litany_2018_CVPR}. This kind of methods works well for objects with a consistent topological structure, such as human faces  \cite{blanz1999morphable,weise2011realtime} and bodies \cite{anguelov2005scape,weiss2011home}. Nevertheless, all these methods deal with every single shape independently, thus cannot transfer the local structural pattern or prior knowledge learned from one shape to another. 

In contrast, learning-based methods leverage deep neural networks to learn shape completion patterns under full supervision on synthetic datasets. These methods mostly adopt an encoder-decoder architecture to extract global feature representations from depth maps \cite{rock2015completing}, RGB images \cite{choy20163d,wu2018learning} discrete SDF voxels \cite{dai2017shape}, or point clouds \cite{stutz2018learning} and subsequently produce a full volumetric shape using the learned priors. Early efforts \cite{wu2016learning,tan2018variational,park2019deepsdf,tan2018variational} mostly use deep neural networks to predict discrete 3D shapes. For example, some early methods \cite{wu2016learning,tan2018variational,park2019deepsdf} explore generative models to generate 3D shapes. In \cite{tan2018variational}, the author build a mesh variational auto-encoders (mesh VAE) to generate 3D mesh from the probabilistic latent space. \cite{litany2018deformable} further enhances \cite{litany2018deformable} by replacing fully connected layers with graph convolutional layers. 
More recent works \cite{park2019deepsdf,mescheder2019occupancy} explore deep neural networks to learn the implicit functions of 3D shapes in the continuous space. In \cite{park2019deepsdf}, the author proposes to use the signed distance function to represent a 3D shape. A generative model is trained to produce a continuous field for 3D shape representation, which is characterized by the zero iso-surface decision boundaries of discretized SDF samples. A similar idea comes from Occupancy Networks \cite{mescheder2019occupancy} that uses the binary voxel occupancy to implicitly represent the 3D surface as the continuous decision boundary of discretized voxel occupancy.

\subsection{Latent Space Optimization}
Instead of using an explicit encoder network for feature learning, one can search the optimal latent representations by training a decoder only network. Tan et al. \cite{tan1995reduc} propose probably the first work to use this idea that simultaneously optimizes the latent codes and the decoder network parameters through back-propagation. During inference, the decoder parameters are fixed as a prior, and a latent code is randomly initialized from a Gaussian distribution and optimized to reconstruct the new observation. Similar approaches have been explored in a large amount of works \cite{rusu2018meta,bouakkaz2012combined,reddy1998input,qunxiong2006dimensionality,groueix20183d}, for applications such as noise reduction, missing measurement completions, fault detection, meta-learning, and shape correspondence. For example, in \cite{groueix20183d}, the author optimizes the latent code extracted from the encoder network to minimize the Chamfer distance between deformed shapes and target shapes, which provides a strong performance boost for shape correspondence. In \cite{park2019deepsdf}, the authors propose to search the optimal latent representations for shape completion. In the paper, we call this type of encoder-free network as auto-decoders. In this paper, both our point set registration network and shape completion network are build upon auto-decoders. 

\section{Method}
In this section, we introduce the proposed method for partial point set registration via \textbf{J}oint shape \textbf{C}ompletion and \textbf{R}egistration using deep neural \textbf{Net}work, called JCRNet. Our method contains two parts: a shape completion network to recover full shapes from corresponding partial observations, and a point set registration network that conducts point set registration from source shapes to target shapes. First, we define the partially registration problem in section \ref{sc_problem_stat}. Then, it gives an overview of the proposed method in section \ref{sc_overview}. We introduce our shape registration network and shape completion network in section \ref{sc_reg} and section \ref{sc_comp} respectively. Finally, we introduce our optimization strategy in section \ref{sc_optim}.

\subsection{Problem Statement}\label{sc_problem_stat}
Given source point sets $\mathcal{X} =  \{\mathcal{X}_1, \mathcal{X}_2, ..., \mathcal{X}_M\}$ and target point sets $\mathcal{Y} =  \{\mathcal{Y}_1, \mathcal{Y}_2$, $..., \mathcal{Y}_M\}$, $\mathcal{X}_i \in \mathbb{R}^{N\times3}, \mathcal{Y}_i \in \mathbb{R}^{N\times3}$, $i \in \{1,2,...,M\}$, our goal here is for each pair of point sets to estimate the rigid transformation from source point set $\mathcal{X}_i$ to target point set $\mathcal{Y}_i$. Without loss of generality, the rigid transformation here is represented by a homogeneous transformation matrix, $p \in SE(3)$, which is composed by a rotation matrix $R \in SO(3)$ and a translation vector $T \in \mathbb{R}^3$, i.e., $p = [R|T]$. 

To properly deal with partial observations, we incorporate a shape completion task along with the point set registration task. For this task, our model takes the partial point set $\mathcal{S}$ as input and outputs a recovered point set $\mathcal{S}'$ to better approximate the full shape. In this paper, we design two separate deep neural networks to learn shape completion function $g: \mathcal{S} \rightarrow \mathcal{S}'$ and the point set registration function $f: \mathcal{X} \rightarrow \mathcal{Y}$.

To integrate these two tasks for joint optimization, we further proposed to incorporate a learnable latent code for each target shape, which can be regarded as a global encoding of the shape. Our method benefits from this design by two points: 1) first, we do not need to explicitly define a feature encoding network to learn the shape representation. Previous methods mostly rely on hand-crafted features or learning deep feature embeddings with deep neural networks. However, neither hand-crafted features nor a learning-based feature encoding network can guarantee the robustness towards partial observations. 2) both our shape completion network and point set registration network takes as input the shared latent code representation, it, therefore, enables the communication of these two tasks and enhances the registration performance. In this way, both our point set registration network and shape completion network benefit from the design of auto-decoder architecture. During training, we optimize all latent codes along with the network parameters. While in the inference stage, we fix the network parameter and optimize these latent codes to get the optimal shape completion and registration results.

\subsection{Method Overview}\label{sc_overview}

\begin{figure*}[h]
    \label{fig_pipeline}
    \centering
    \includegraphics[width=18cm]{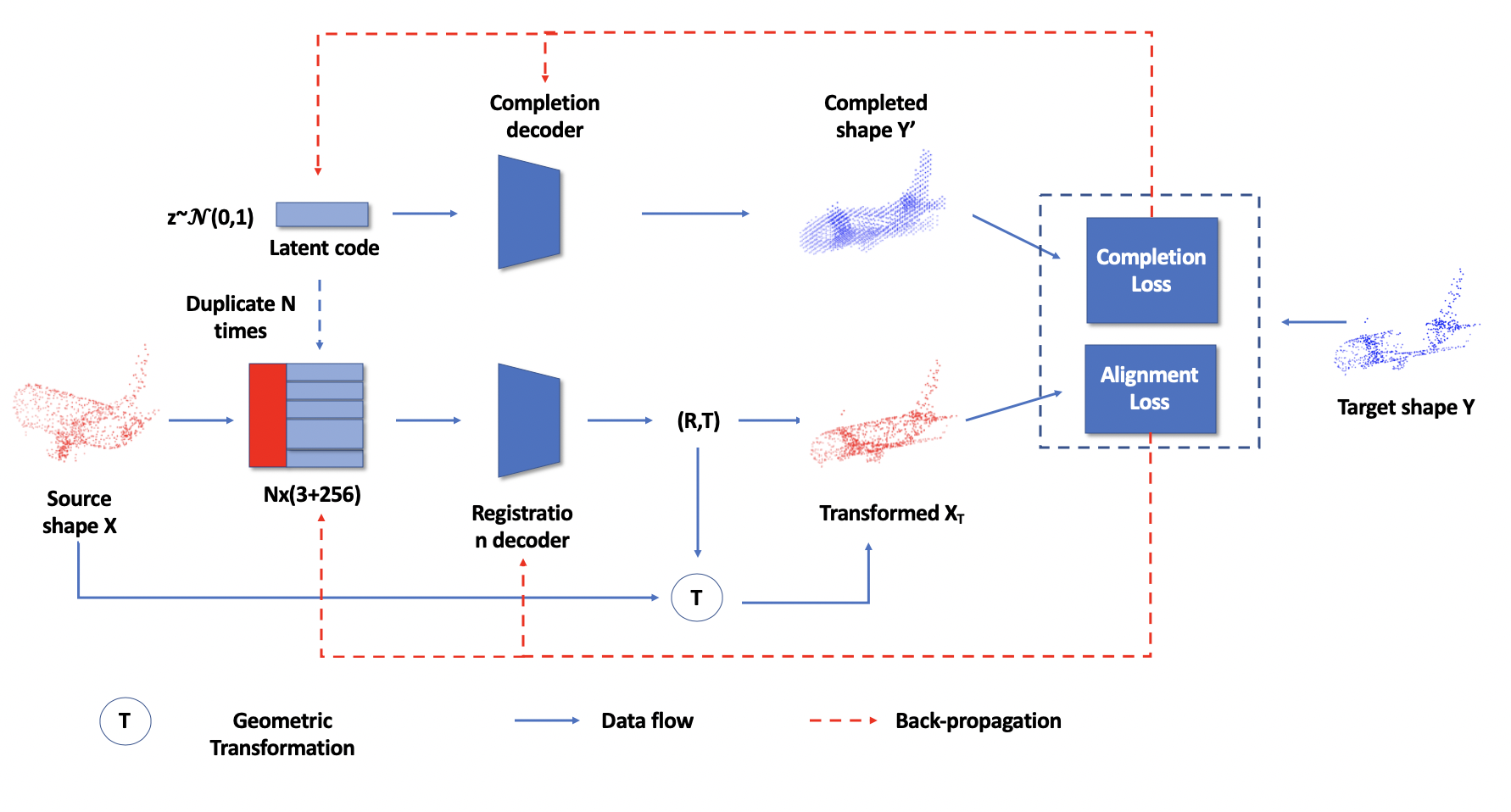}
    \caption{Method overview. Our method contains two parts: a shape completion network to recover full shapes from corresponding partial observations (upper part), and a point set registration network conduct point set registration from the recovered shapes (lower part). Both parts use an auto-decoder network that takes as input a randomly initialized latent code of target shape and optimizes the latent code during network training. 
More specifically, given a pair of input point sets, source point sets $\mathcal{X} \in \mathbb{R}^{N\times3}$ and target point set $\mathcal{Y} \in \mathbb{R}^{N\times3}$, our model incorporates a learnable latent code $z  \in \mathbb{R}^{256}$ for each target point set, which can be regarded the global representation of target point set. The shape completion network takes the latent code $z$ of target shape $\mathcal{Y}$ as inputs and produces the recovered shape $\mathcal{Y}'$ under the supervision of $\mathcal{Y}$. In this paper, we use an auto-decoder network to reconstruct the partial target shape $\mathcal{Y}$ following \cite{park2019deepsdf} from a learnable latent code. The point set registration network takes the source point sets $\mathcal{X}$ and the latent code $z$ as input and predicts the rigid transformation from source point set $\mathcal{X}$ to target point set $\mathcal{Y}$. More specifically, we stack the coordinates of each point on shape $\mathcal{X}$ with the latent code $z$ to formulate the inputs for our point set registration network. We use a fully connected network to regress the rigid transformation parameters. Our model achieves simultaneous shape completion and registration from partial observations.
}
\label{fig_pipeline}
\end{figure*}

Fig. \ref{fig_pipeline} gives an overview of the proposed method for partial point set registration. Our method contains two parts, and both parts use an auto-decoder architecture. In the first part (upper part in Fig. \ref{fig_pipeline}), our method performs shape completion for the target shape $\mathcal{Y}$. It takes the latent code $z$ of the target shape as input. Then we design a decoder network $g_{\beta}$ to recover the full point set with the partial observation as supervision. In the second part (lower part in Fig. \ref{fig_pipeline}), our model learns the geometric transformation from source point set $\mathcal{X}$ to the target point set $\mathcal{Y}$. Specifically, we stack the coordinates of each point in source shape with latent vector $z$ as the input, i.e., $[z,x]$, for each $x \in \mathcal{X}$, and feed them to another decoder network $f_{\theta}$ to predict the rigid transformation. By applying the predicted geometric transformation to source shape, we then formulate an unsupervised alignment loss between the transformed source point set $\mathcal{X}_{\mathcal{T}}$ with the source point set $\mathcal{Y}$. The shared latent codes are initialized from a Gaussian distribution. During training, the latent codes are optimized along with two decoder networks. While in the inference stage, we fix the network parameters and optimize the latent code for each pair of input shapes to produce the optimal geometric transformation.

\subsection{Global Feature Representation}\label{sc_scr}
To characterize the global feature of a given point set, previous methods need to explicitly define a feature encoder network (i.e., PointNet \cite{qi2017pointnet}) for the deep spatial feature extraction. However, the design of an appropriate feature encoder for unstructured point clouds is challenging as standard discrete convolutions assume the availability of a structured input (e.g., 2D image). To avoid the hand-crafted network design for the extraction of spatial features, in this paper, we define a learnable latent code for each partial target point set with an intention to characterize the essence of its global feature. As shown in Fig. \ref{fig_pipeline}, the latent code is randomly initialized from a Gaussian distribution and is fed into two different decoder networks for the dual tasks of shape completion and registration, respectively. 

By the design of this latent code representation, both our shape completion network and shape registration network takes as input the same global feature representation and is optimized along with the network parameters during training. In the inference stage, we fix the network parameters and only optimize the latent code for each pair of input shapes. In this way, the latent codes are enforced to characterize the global feature of a 'full shape' instead of a partial one. 

In this paper, we use auto-decoder architecture for two reasons: 1) previous research \cite{park2019deepsdf} has shown that using auto-decoder architecture can lead to better performance compared with encoder-decoder architecture during SDF learning; 2) use auto-decoder architecture enables a joint optimization of our shape completion and point set registration tasks.

\begin{figure*}[h]
    \centering
    \includegraphics[width=17cm]{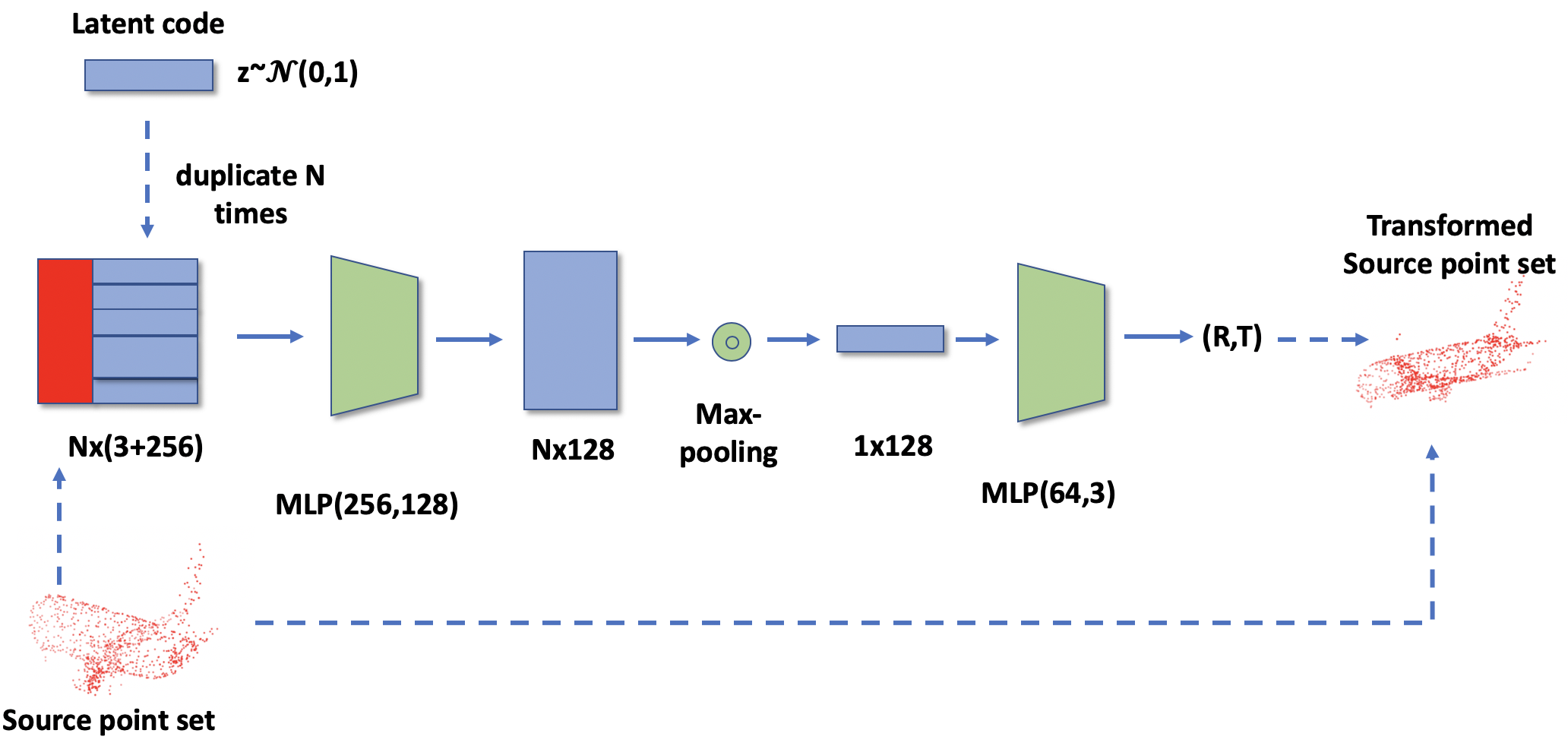}
    \caption{Architecture of our point set registration network. Given a pair of source and target point set with dimension of $N\times 3$, and the associating latent code with a dimension of 256, our model predicts the desired geometric transformation (R,T).}
    \label{fig_reg}
\end{figure*}

\subsection{Point Set Registration}\label{sc_reg}

Our point set registration network $f_{\theta}$ takes as input the source point set $\mathcal{X}_i$ and the latent vector $z_i$ of the target point set $\mathcal{Y}_i$, and it outputs the rigid transformation from source point set $\mathcal{X}_i$ to target point set $\mathcal{Y}_i$. The architecture of our point set registration network is illustrated in Fig. \ref{fig_reg}. For each point on shape $\mathcal{X}_i$, we stack its coordinates with the associating latent vector to formulate the input $[z_i, \mathcal{X}_i]$ and feed it into transformation decoder network $f_{\theta}$ to predict the desired transformation.

In this paper, our transformation decoder network firstly uses a multi-layer perceptron network to map the inputs to high-dimensional feature embeddings, followed by a max-pooling layer to get the global transformation representation. Several fully connected layers are further used to map the global transformation representation to desired geometric transformation, i.e., three rotation angles $r_i = (rx, ry, rz)$ and a translation vector $t_i=(ta,tb,tc)$.
\begin{equation}
    [r_i,T_i] = f_{\theta}([z_i, \mathcal{X}_i])\\
\end{equation}
The rotation matrix $R_i$ can then be formulated from the ration angles as,
\begin{equation}
    R_i = Rz \odot Ry \odot Rx\\
\end{equation}
\begin{equation}
    Rz = \begin{pmatrix}
    cos(rz) & -sin(rz) & 0\\
    sin(rz) & cos(rz) & 0\\
    0 & 0 & 1\\
    \end{pmatrix}\\
\end{equation}
\begin{equation}
    Ry =\begin{pmatrix}
    cos(ry) & 0 & sin(ry)\\
    0 & 1 & \\
    -sin(ry) & cos(ry) & 0 \\
    \end{pmatrix}\\
\end{equation}
\begin{equation}
    Rx =\begin{pmatrix}
    1 & 0 & 1\\
    0 & cos(rx) & -sin(rx)\\
    0 & sin(rx) & cos(rx)\\
    \end{pmatrix}
\end{equation}
where $\odot$ denotes matrix multiplication.

The transformed source point set can be generated by,
\begin{equation}
\mathcal{X_T}_i = R_i\mathcal{X}_i+t_i\\
\end{equation}

Generally, Chamfer distance can be used to measure the alignment between two point sets. In this paper, considering the target point set is a partial observation, we use the clipped Chamfer distance to measure the alignment loss between transformed source shape $\mathcal{X_T}_i$ and partial target shape $\mathcal{Y}_i$, i.e., we penalize the distance from each point on partial target shape $\mathcal{Y}_i$ to the nearest point on transformed source shape $\mathcal{X_T}_i$ and also the distance from each point on transformed source shape $\mathcal{X_T}_i$ to its nearest point on target shape $\mathcal{Y}_i$. Our registration loss is calculated as follows,

\begin{equation} 
\begin{split}\mathcal{L}_{reg}(\mathcal{X_T}_i, \mathcal{Y}_i) &= \sum_{y\in \mathcal{Y}_i} \text{min}(\sigma_t, \min_{x \in \mathcal{X_T}_i}||x-y||^2_2)\\
&+ \sum_{x \in \mathcal{X_T}_i} \text{min}(\sigma_t, \min_{y\in \mathcal{Y}_i}||x-y||^2_2)\\
\end{split}
\label{eq_reg}
\end{equation}
where $\sigma_t$ denotes the threshold for clipping the distance.

In the training stage, we simultaneous optimize registration decoder network parameters $\theta$ and the latent vector $z_i$ using all shape pairs in the training set. While in the inference stage, we fix the decoder network parameter and optimize the latent vector for each input pair of point set to produce the desired rigid transformation.

\begin{figure*}[!h]
    \centering
    \includegraphics[width=17cm]{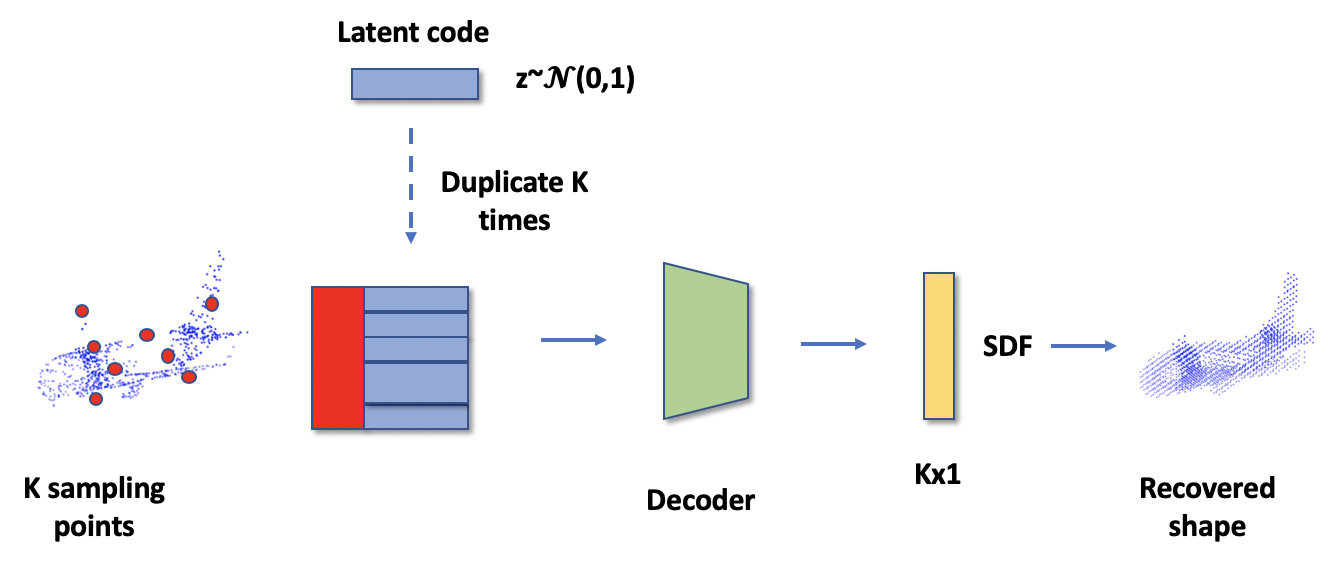}
    \caption{Architecture of our shape completion network. Given a partial target point set and its associating latent code, our network produces $K$ SDF values for $K$ sampled spatial locations. }
    \label{fig_comp}
\end{figure*}

\subsection{Shape Completion}\label{sc_comp}
In this paper, we design the shape completion network to recover the full shapes from partial observations. It enforces the latent code to represent the global representation for the full shapes instead of partial ones and thus also makes it easier for the dual task of point set registration. 

Our shape completion network is designed using auto-decoder architecture following the method presented in DeepSDF \cite{park2019deepsdf}. To represent a shape $S$, we aim to learn a signed distance function (SDF) that, for any given spatial point $p_{ij}$, it outputs the distance from this point to the closest point on the surface of the shape.
$S$, $SDF(S,p_{ij}) = s_{ij}$. Note that $p_{ij}$ does not need to be on shape $S$. The sign of SDF value indicates whether the point is inside (negative) or outside (positive) the input surface. We refer interested readers to \cite{park2019deepsdf} for more details.

More specifically, given each target shape $\mathcal{Y}_i$, a latent code $z_i$ is paired with the shape to represent its shape information. The SDF value at spatial location $y_{ij}$ can be generated via a deep feed-forward network $g_{\beta}(\cdot) $, see Eq. (\ref{eq_sdf}). Fig. \ref{fig_comp} illustrates the pipeline of our shape completion network.

\begin{equation}
SDF(s_{ij}|z_i, y_{ij}) = g_{\beta}(z_i, y_{ij} )\\
\label{eq_sdf}
\end{equation}

In the training phase, our model tries to minimize a pre-defined loss function over all training shapes with respect to all latent codes $\{z_i\}_{i=1}^{M}$ and the network parameters $\theta$:
\begin{equation} 
\mathcal{L}_{com}(z_i, \mathcal{Y}_i) = \frac{1}{M} \sum_{i=1}^{M} \sum_{j=1}^{K} (\mathcal{L}(f_{\theta}(z_i, y_{ij}), s_{ij}) + \frac{1}{\sigma^2}||z_i||_2^2)
\end{equation}
where $M$ denotes the number of training shapes, $K$ denotes the number of point samples for each shape. $\mathcal{L}(\cdot, \cdot)$ denotes the loss function penalizing the predicted SDF value $s_{ij}'$ and ground truth SDF value $s_{ij}$. The second term is used to penalize the latent codes. $\sigma$ is the hyperparameter to balance these two loss terms. We use the clamped $L1$ loss for $\mathcal{L}$ with a threshold of 0.03. The point sampling strategy will be explained in Section \ref{sc_details}. At inference time, we fix the network parameters $\beta$, thus a shape code $z_i$ for each shape $\mathcal{Y}_i$ can be estimated via Maximum-a-Posterior (MAP) estimation. 

\subsection{Optimization Strategy}\label{sc_optim}
We train our shape completion and point set registration network with the full shapes in training set. In the training stage, we simultaneously optimize the parameters of our shape completion network and point set registration network, as well as the latent codes for all shapes. The optimal network parameters can be generated follows
\begin{equation}
\theta^{optim}, \beta^{optim} = \mathop{\arg \min}_{\theta, \beta, \{z_i\}_{i=1}^{M}}   (\mathcal{L}_{reg}([z_i, \mathcal{X}_i], \mathcal{Y}_i)+\lambda \mathcal{L}_{com}(z_i, \mathcal{Y}_i) )
\label{eq_opt_train}
\end{equation}
where the first term denotes point set registration loss and the second term denotes shape completion loss, and $\lambda$ denotes the hyper-parameters to balance these two tasks. Without specific mention, we set $\lambda$ to 0.1 in the following sections.
After training on full shapes from the training set, the learned decoder network parameters $\theta$ and $\beta$ here provides a prior knowledge for the optimization of latent code representation. 

During model evaluation, given a pair of partial observation with source point set $\mathcal{X}_i$ and target point set $\mathcal{Y}_i$, we fix the two decoder networks and only optimize the latent code of partial target shape $\mathcal{Y}_i$. To achieve this, we randomly initialize a latent code $z_i$ for the target point set and optimize it via Maximum-a-Posterior (MAP) estimation as:
\begin{equation}
z_i^{optim} = \mathop{\arg \min}_{z_i}   (\mathcal{L}_{reg}(f_\theta([z_i, \mathcal{X}_i]), \mathcal{Y}_i) + \lambda 
\mathcal{L}_{com}(z_i, \mathcal{Y}_i))
\label{eq_opt_test}
\end{equation}
After the optimization process, the optimal latent code can be used to perform shape completion for the target shape as well as predict the desired geometric transformation from source point set to target point set simultaneously using equation (1) and equation (4). One may note that it would be possible to split our model into a two-step process: 1) optimize latent code $z_i$ to get optimal target shape reconstruction, 2) optimize latent code $z_i$ using the Chamfer distance between the reconstructed target shape $\mathcal{Y}_i'$ and transformed source shape $\mathcal{X_T}$. However, in our experiments (in section \ref{sc_exp_ab}), we find that this two-step process does not lead to better performance compared to our model using end-to-end training. 

\begin{figure*}[!h]
    \centering
    \includegraphics[width=17cm]{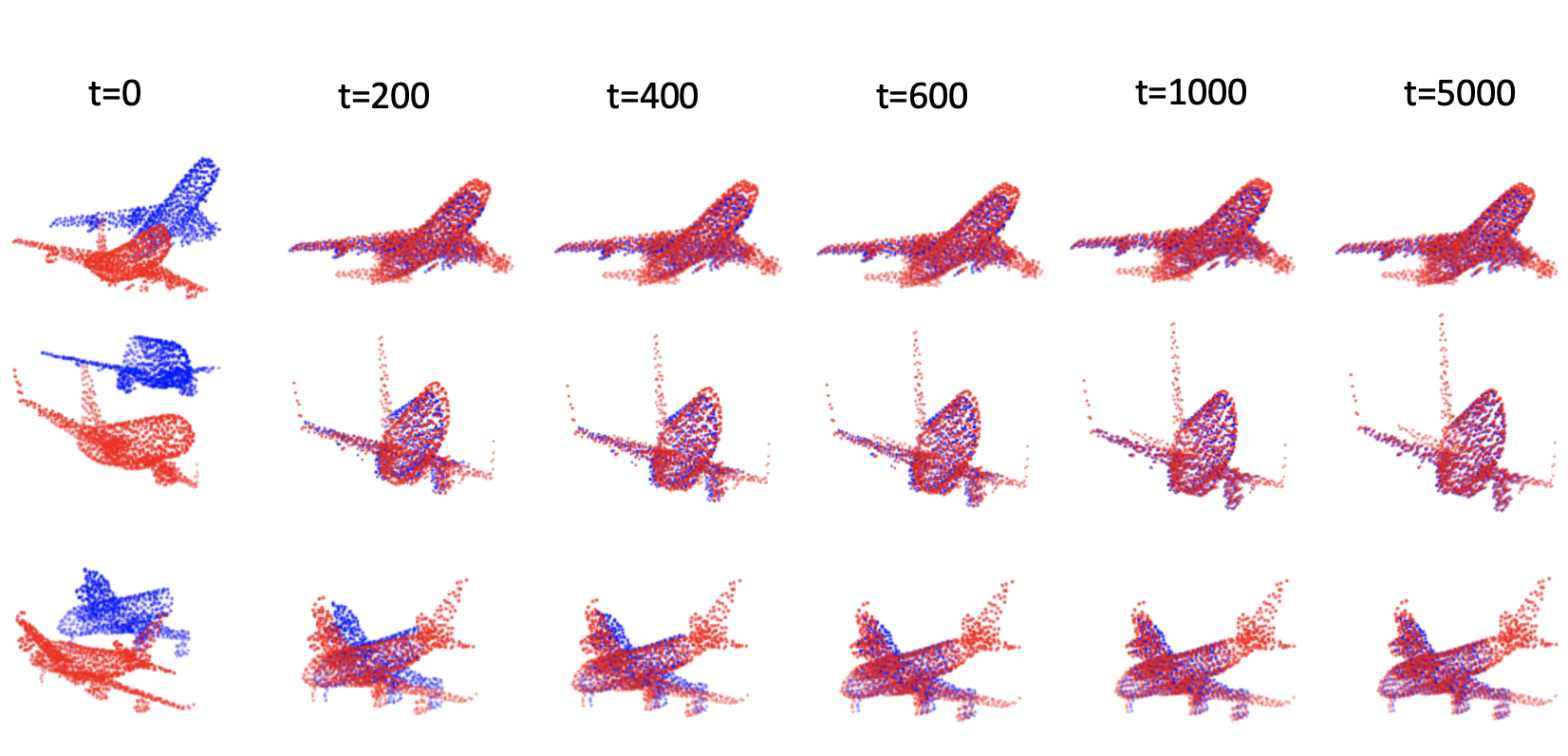}
    \caption{Illustration of alignment process. The red points represent source point sets and the blue points represent the target point sets. We show the alignment results after 0, 200, 400, 600, 1000 and 5000 steps.}
    \label{fig_proc}
\end{figure*}
\section{Experiments}
\subsection{Experimental Dataset}
We evaluate the proposed JCRNet for partially registration on ModelNet40 \cite{wu20153d} benchmark dataset. ModelNet40 dataset contains 12,311 shapes of 40 object categories. We split this dataset into  9,843 for training and 2,468 for testing. We train our model using full shapes from the training set and evaluate the registration performance with the partially shapes from the test set. 

To generate the data for training and model evaluation, we randomly sample 1024 points on each shape. Then, for each source shape $\mathcal{X}$ we generate the transformed shapes $\mathcal{Y}$ by apply a rigid transformation: the rotation matrix is characterized by three rotation angles along the xyz axes, where each value is uniformly sampled from $[0,45]$ unit degree, and the translation is uniformly sampled from $[-0.5, 0.5]$. At last, we simulate partial point sets of $\mathcal{Y}$ by randomly select a point in unit space and keep its 768 nearest neighbors.

\subsection{Implementation details}\label{sc_details}
For the point set registration network, we use the sequential structure of C(256)-C(128)-M-FC(128)-FC(64)-FC(3), where C denotes 1D convolution layer with a kernel size of 1, FC denotes fully-connected layer, and M denotes 1D max-pooling. To formulate a rigid transformation, our point set registration network predicts a 3-dimensional output for rotation angles and translation vector. 

For the shape completion network, we leverage the same architecture as \cite{park2019deepsdf} and use a 7-layer MLP to decode the latent code into SDF values, each layer with a size of 512. To train the completion network, we need to construct numerous of SDF samples for each target shape. Each SDF sample include a 3D point and its SDF value. Specifically, we perturb each point on the target shape with zero-mean Gaussian noise with standard deviation of 0.2 to generate three spatial samples per point. For each point sample, we find the nearest point from the original target shape and calculate the distance as its SDF values.

Our model is optimized using Adam optimizer with an initial learning rate of 1e-3. We set the batch size to 50, momentum to 0.9, and weight decay to 1e-5. We use batch normalization and dropout for all fully connected layers, except for the output layer, both in our shape completion network and in point set registration network. 
The latent vectors $\{z_i\}$ are randomly initialized from a Gaussian distribution $\mathcal{N}(0, 0.06)$ with a dimension of 256. Our method is implemented with the PyTorch library on a TESLA K80 GPU.

We evaluate the registration performance by the metrics, including mean squared error (MSE), root mean squared error (RMSE), and mean absolute error (MAE). Angular measurements are in units of degrees. To demonstrate the effectiveness of our proposed JCRNet, we compare our model to Closest Point (DCP) \cite{wang2019deep} and PR-Net \cite{wang2019prnet}. We omit other comparing methods because DCP and PR-Net have demonstrated state-of-the-art point set registration performance for full observations and partial observations respectively.


\subsection{Transformation Process}
Before going to the experimental results, we examine the alignment process of three test pairs by the optimization steps. As shown in Fig. \ref{fig_proc}, our proposed model is able to align the main parts of the input shapes after 100 optimization steps. The rotation predictions are close to the ground truth ones, while it still takes thousands of optimization steps for our model to get the accurate translation predictions. After 5000 steps, our model produces satisfying alignment for all input shape pairs.

\begin{figure*}[t]
    \centering
    \includegraphics[width=18cm]{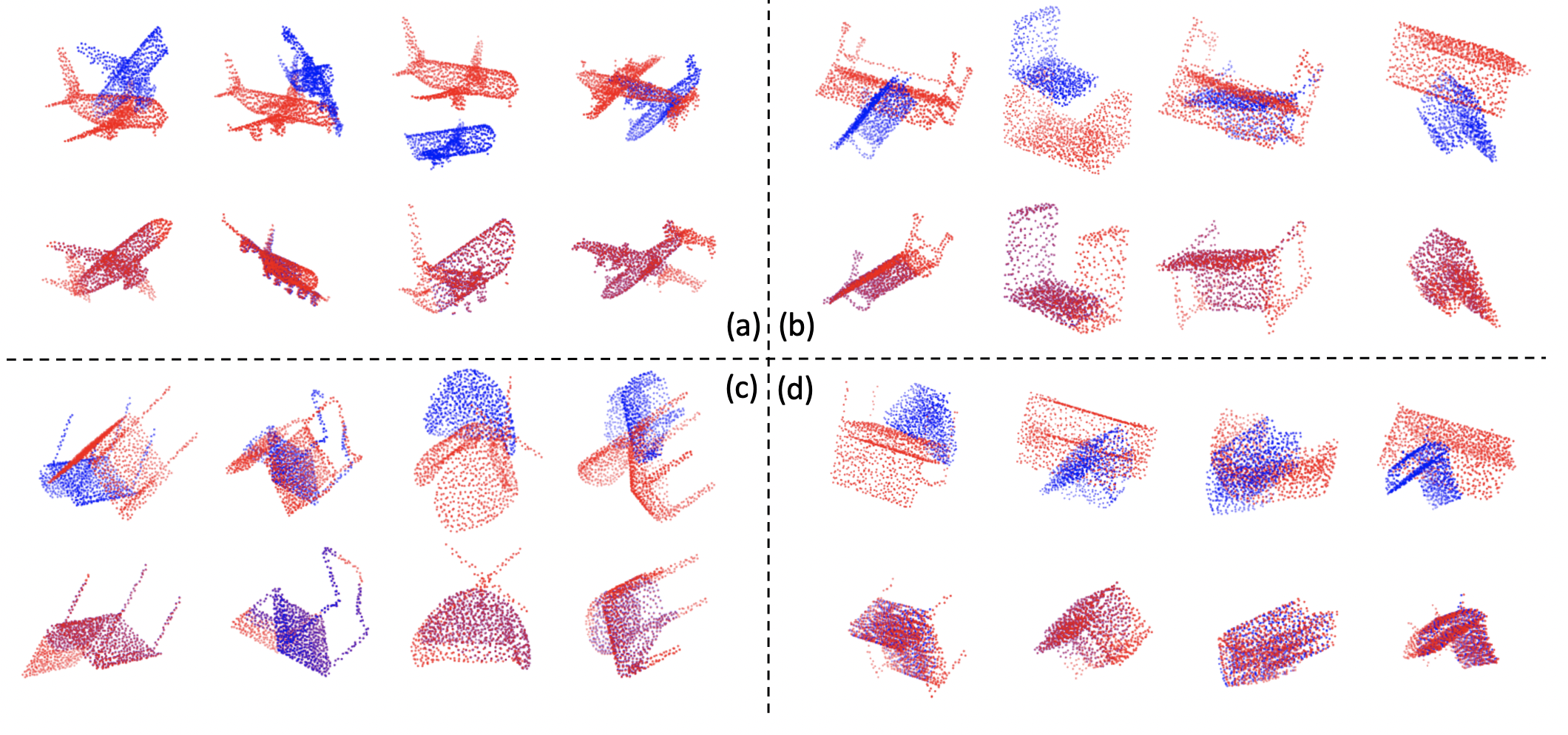}
    \caption{Randomly selected examples of partial point set registration on different categories, (a) airplane, (b) bench, (c) chair and (d) sofa. The red points represent source point sets and the blue points represent the target point sets. The odd rows show input shapes, and the even columns show output results.}
    \label{fig_train_test}
\end{figure*}

\begin{table*}[t]
    \centering
    \begin{tabular}{l r r r r r r}
        \hline
        Model & MSE(R) & RMSE(R) & MAE(R) & MSE(t) & RMSE(t) & MAE(t) \\
        \hline
        DCP \cite{wang2019deep} & 20.518583 & 4.529744 & 3.281942 & 0.001546 & 0.039315 & 0.029034\\
        PRNet \cite{wang2019prnet} & 6.349841 & 2.519889 & 1.926538 & 0.002804 & 0.052949 & 0.041235\\
        JCRNet (Ours) & 0.059714 & 0.234741 & 0.173260 & 0.000467 & 0.021502 & 0.017629\\
        \hline
        \hline
      DCP \cite{wang2019deep} & 13.646673 & 3.694140 & 2.743233 & 0.002339 & 0.048364 & 0.036228\\
        PRNet \cite{wang2019prnet} & 9.687140 & 3.112417 & 2.276538 & 0.003241 & 0.056932 & 0.044811\\
      JCRNet (Ours) & 0.049004 & 0.214932 & 0.151105 & 0.001628 & 0.040348 & 0.010773\\
        \hline
        \hline
        DCP \cite{wang2019deep} & 35.288406 & 5.940404 & 4.494213 & 0.001378 & 0.037116 & 0.029175\\
        PRNet \cite{wang2019prnet} & 13.405939 & 3.661412 & 2.761630 & 0.001313 & 0.036239 & 0.029226\\
         JCRNet (Ours) & 0.214308 & 0.266104 & 0.123464 & 0.000074 & 0.007140 & 0.003060\\
        \hline
        \hline
        DCP \cite{wang2019deep} & 10.668366 & 3.266247 & 2.458252 & 0.000892 & 0.029860 & 0.022282\\
        PRNet \cite{wang2019prnet} & 5.853882 & 2.419480 & 1.819235 & 0.001110 & 0.033323 & 0.025116 \\
        JCRNet (Ours) & 0.179997 & 0.408753 & 0.287125 & 0.001675 & 0.039729 & 0.018096\\
        \hline
    \end{tabular}
    \caption{From top to bottom: test performance on airplane, bench, chair and sofa category respectively.}
    \label{tab_unseen_obj}
\end{table*}

\subsection{Results on ModelNet40}\label{sc_exp_results}
We first evaluate the performance of our JCRNet on the ModelNet40 dataset. Table \ref{tab_unseen_obj} listed the performance of our method and the comparing methods, including DCP and PR-Net on the airplane, bench, chair categories of the ModelNet40 dataset. As shown in Table \ref{tab_unseen_obj}, our model gets an order of magnitude better performance than the comparing methods for the rotation prediction and also achieves significantly better performance for translation prediction. Specifically, for the rotation prediction, our JCRNet achieves an RMSE(R) less than 1 degree for all four classes. 
Note that both DCP and PR-Net solve the geometric transformation based on dense point correspondences between the transformed source shape and the target shapes. Nevertheless, it's still a non-trivial to find the dense point correspondences using point-wise features, especially for partial observations. In contrast, our JCRNet model does not use point correspondence information but directly regress the geometric transformation using fully connected layers. This makes our model can be trained in a purely unsupervised way and also makes it more flexible to be equipped in different network architectures.
Some randomly selected examples are shown in Fig \ref{fig_train_test}. As shown in Fig \ref{fig_train_test}, our model is able to align partial inputs with different orientations on all four categories. We do not include the results of other categories here because our model is trained independently for each category. It would take much time to conduct experiments on all 40 categories of the ModelNet40 dataset.

\subsection{Resilience to Gaussian Noise}
\begin{figure*}[h]
    \centering
    \includegraphics[width=18cm]{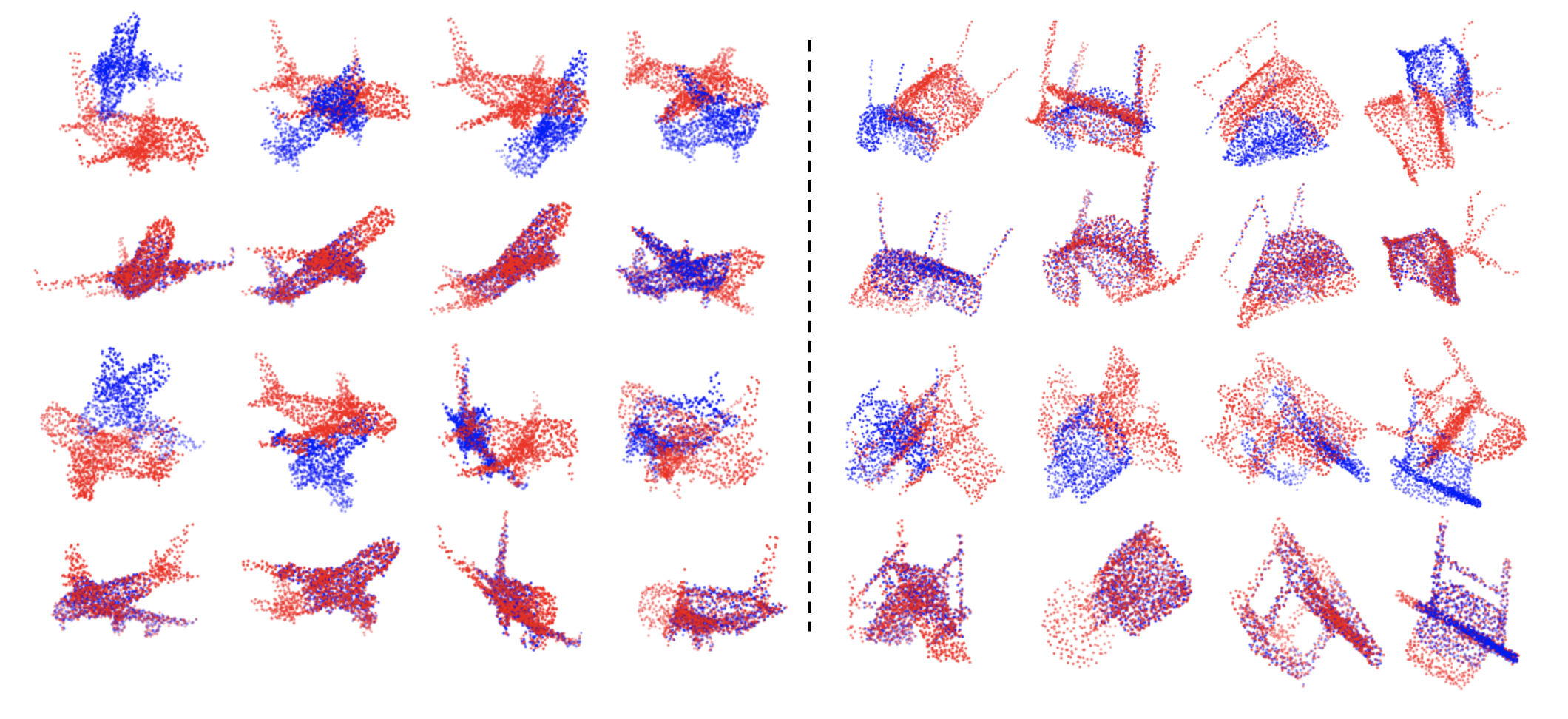}
    \caption{Visualization of the registration results with Gaussian noise on airplane (left) and chair (right) categories. $N$ indicates the remaining number of points in the partial target shape. The odd rows show input shapes and the even rows show output results.}
    \label{fig_gauss}
\end{figure*}

We further test the robustness of our model under Gaussian noise. To do this, we add noise to each of the points on both source and target point sets, which was randomly sampled from $\mathcal{N}(0,0.01)$ and clipped to $[-0.05, 0.05]$. Other settings are same as section \ref{sc_details}. We trained our JCRNet model and comparing methods, including DCP and PR-Net, on noise-free data and evaluate the performance of all methods on the test set with Gaussian noise. Evaluation results are listed in Table \ref{tab_gaussian}. As shown in this table, our model achieves significantly better performance than the comparing methods for both rotation and translation prediction. By comparing Table \ref{tab_unseen_obj} and Table \ref{tab_gaussian}, one can find that on the sofa category, the performance of the DCP model degrades a lot, with RMSE(R) increase from 3.266247 to 3.864215. The performance of our model and PR-Net \cite{wang2019prnet} decreases much smaller, with RMSE(R) increases from 2.419480 and 0.408753 to 2.497723 and 0.484324, respectively. This demonstrates that comparing to the DCP model, our model and PR-Net are more robust to data noise for the task of partial point set registration. Moreover, compared with PR-Net, our JCRNet model gets a smaller performance drop. Fig. \ref{fig_gauss} shows some selected examples of our model on airplane and chair categories. As shown in Figure \ref{fig_gauss}, even when the input point sets are with Gaussian noise, our model can still successfully align the source and target point sets.

\begin{table*}[]
    \centering
    \begin{tabular}{l r r r r r r}
        \hline
        Model & MSE(R) & RMSE(R) & MAE(R) & MSE(t) & RMSE(t) & MAE(t) \\
        \hline
        DCP \cite{wang2019deep} & 27.383821 & 5.232955 & 3.716614 & 0.002059 & 0.045373 & 0.034398\\
        PR-Net \cite{wang2019prnet} & 7.609180 & 2.758474 & 2.119697 & 0.003375 & 0.058099 & 0.045531\\
        JCRNet (Ours) & 3.540532 & 1.273092 & 0.376848 & 0.000425 & 0.017153 & 0.005731\\
        \hline
        \hline
        DCP \cite{wang2019deep} & 18.817074 & 4.337865 & 3.29574 & 0.004637 & 0.068093 & 0.049714\\
        PR-Net \cite{wang2019prnet} &9.868777 & 3.141461 & 2.333664 & 0.003426 & 0.058536 & 0.045699\\
        JCRNet (Ours) & 0.058920 & 0.228857 & 0.138904 & 0.003276 & 0.057238 & 0.019733\\
        \hline
        \hline
        DCP \cite{wang2019deep} & 48.763847 & 6.983111 & 5.391983 & 0.00156 & 0.039496 & 0.031935\\
        PR-Net \cite{wang2019prnet} &13.439675 & 3.666016 & 2.746882 & 0.001359 & 0.036871 & 0.030141\\
        JCRNet (Ours) & 0.305641 & 0.406611 & 0.198128 & 0.000040 & 0.005002 & 0.002327\\
        \hline
        \hline
        DCP \cite{wang2019deep} & 14.932157 & 3.864215 & 2.872617 & 0.001026 & 0.032031 & 0.024701\\
        PR-Net \cite{wang2019prnet} & 6.238619 & 2.497723 & 1.836021 & 0.001127 & 0.033567 & 0.025541\\
        JCRNet (Ours) & 0.261543 & 0.484324 & 0.331687 & 0.002196 & 0.044315 & 0.024815\\
        \hline
    \end{tabular}
    \caption{Fro top to bottom: test on unseen point clouds with Gaussian noise on airplane, bench, chair and sofa categories.}
    \label{tab_gaussian}
\end{table*}

\begin{figure*}[h]
    \centering
    \includegraphics[width=18cm]{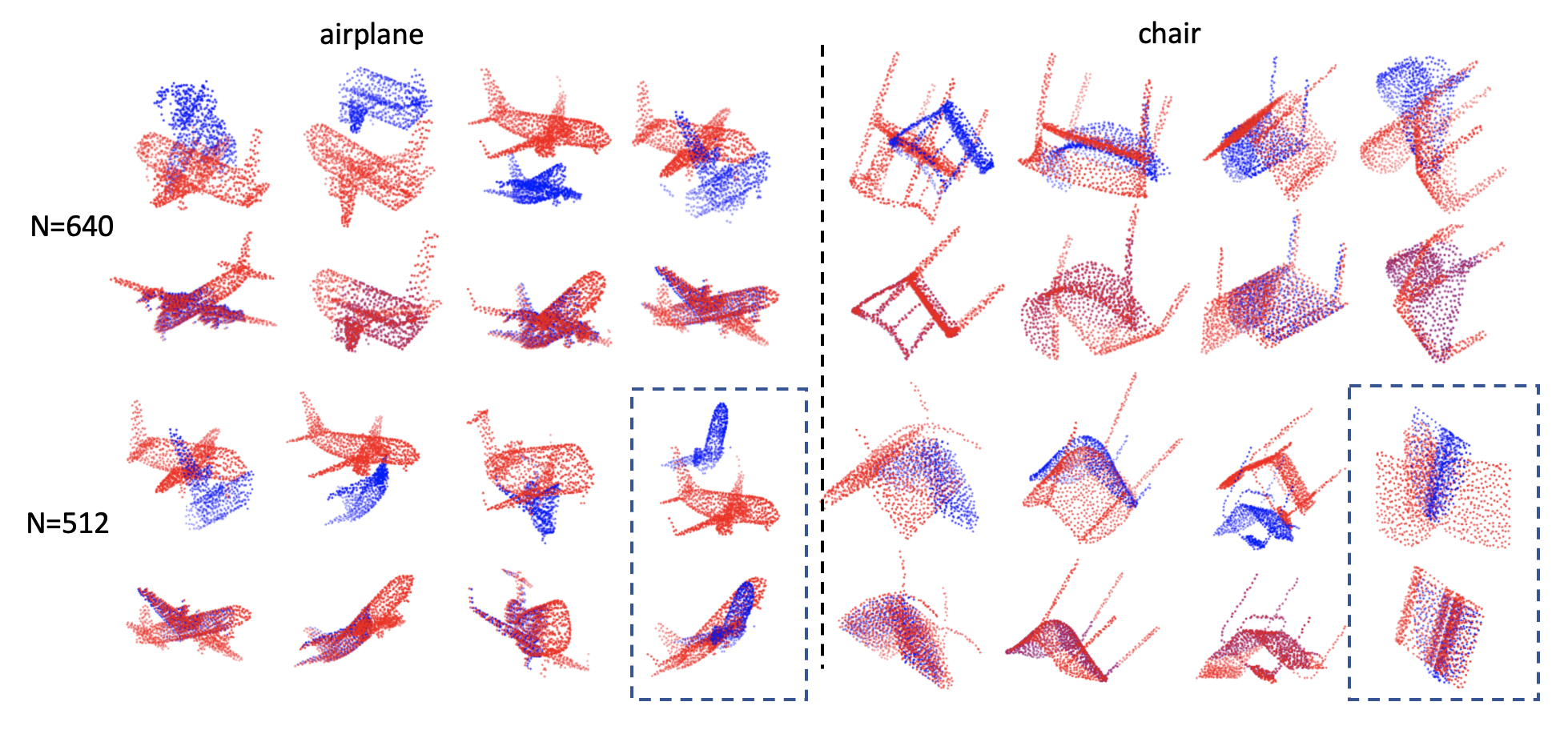}
    \caption{Visualization of different number of missing points on airplane (left) and chair (right) categories. The odd rows show input shapes, and the even columns show output results. Blue  boxes show failure cases.}
    \label{fig_num_missing}
\end{figure*}


\subsection{Number of Missing Points}
In this section, we explore the performance of our model with different numbers of missing points. We conduct experiments on the airplane and chair categories with 256, 384, and 512 missing points, i.e., each partial shape has 768, 640, and 512 remaining points. Table \ref{tab_num_missing} shows the quantitative results of our model. As can be seen in Table \ref{tab_num_missing}, with the increase in the number of missing points, the performance of our model decreases as expected. For the partial shapes with 640 remaining points, our model gets an RMSE(R) around 1 degree and RMSE(t) less than 0.06. Even with 512 missing points, our model can still maintain an RMSE(R) less than 5 degrees for both airplane and chair categories. We also notice that the MSE(R) becomes large when the input shapes have 512 missing points. In our experiments, we notice that our model fails to align several cases, and the rotation predictions are with large errors. This is because when input shapes are with a large number of missing points, our model may converge to local minimal, considering the latent codes $\{z_i\}$ are randomly initialized from Gaussian distribution. These failure predictions dominate the prediction errors, especially for MSE(R). 

Figure \ref{fig_num_missing} shows some selected examples of the registration results of our model on airplane and chair categories with 640 and 512 remaining points. As shown in Fig. \ref{fig_num_missing}, our model can produce satisfying alignment for most of the cases. The cases in the blue boxes show failure cases. For these failure cases, our model predicts the rotation matrix with large errors. In our experiments, we find our model fails for only a few cases for each object category.

\begin{table*}[]
    \centering
    \begin{tabular}{l r r r r r r}
        \hline
        \#points & MSE(R) & RMSE(R) & MAE(R) & MSE(t) & RMSE(t) & MAE(t) \\
        \hline
        768 & 0.058920 & 0.228857 & 0.138904 & 0.003276 & 0.057238 & 0.019733\\
        640 & 1.262889 & 0.992460 & 0.577019 & 0.000553 & 0.023497 & 0.019172\\
        512 & 19.755819 & 3.851332 & 1.201197 & 0.003392 & 0.057192 & 0.021514\\
        \hline
        \hline
        768 & 0.214308 & 0.266104 & 0.123464 & 0.000074 & 0.007140 & 0.003060\\
        640 & 3.638625 & 1.242836 & 0.452099 & 0.000169 & 0.012908 & 0.005759\\
        512 & 20.928379 & 4.574755 & 0.994808 & 0.060251 & 0.245461 & 0.204531\\
        \hline
    \end{tabular}
    \caption{Test performance with different number of missing point on airplane (top) and chair (bottom) categories. `\#points' indicates the remaining number of points in the partial target shape.}
    \label{tab_num_missing}
\end{table*}

\begin{table*}[]
    \centering
    \begin{tabular}{l r r r r r r}
        \hline
        Model & MSE(R) & RMSE(R) & MAE(R) & MSE(t) & RMSE(t) & MAE(t) \\
        \hline
        DCP \cite{wang2019deep} & 20.518583 & 4.529744 & 3.281942 & 0.001546 & 0.039315 & 0.029034\\
        PRNet \cite{wang2019prnet} & 6.349841 & 2.519889 & 1.926538 & 0.002804 & 0.052949 & 0.041235\\
        \hline
        Ours (baseline) & 2.202928 & 1.394375 & 0.522311 & 0.003366 & 0.057668 & 0.024253\\
        Ours ($\lambda=0.1$) & 0.058920 & 0.228857 & 0.138904 & 0.003276 & 0.057238 & 0.019733\\
        \hline
    \end{tabular}
    \caption{Test performance on airplane category of ModelNet40 dataset with and without shape completion for partial observation. The baseline is our model without the shape completion network.}
    \label{tab_ab}
\end{table*}

\subsection{Ablation Analysis}\label{sc_exp_ab}
In this section, we conduct further experiments to demonstrate the effectiveness of our shape completion module. For this, we remove the shape completion part from our method, i.e., set $\lambda$ to 0. In this way, our method directly performs point set registration for the partial point sets. Table \ref{tab_ab} lists the performance of our model with and without shape completion module, as well as the comparing methods, i.e., DCP \cite{wang2019deep} and PR-Net \cite{wang2019prnet}. As can be seen in the table, both our model and our baseline model achieve significantly better performance than the comparing methods. More importantly, our model enjoys a significant performance boost by the joint training of the shape completion network, especially for rotation prediction.

\section{Conclusion}

In this paper, we introduce an unsupervised method for partial point set registration. Given the fact that recent learning-based methods have achieved significantly better registration performance on the full shapes than partial observations, we propose to bridge the performance gaps by incorporating a shape completion network to recover full shapes from partial observations. To achieve this, we design a latent code for each of the target shape, which can be regarded as the global feature encoding the target shape. This latent code is initialized from a Gaussian distribution and is taken as the inputs for both shape completion and registration networks. In this way, our model eliminates the explicit design of point feature encoding network and, more importantly, enables the joint training of both shape completion and registration networks and therefore boosts the performance of our registration network. During training, the latent code is optimized along with the parameters of shape completion and registration networks. While in the inference stage, the network parameters are fixed and used as prior knowledge to guide the optimization of the latent codes to get the optimal shape completion and registration results. Experiments on the ModelNet40 dataset demonstrate the effectiveness of our model for rigid transformation prediction on partial observations. Results also show that our model is robust to inputs with Gaussian noise.

\section{Acknowledgement}
This work was supported by the NYU Abu Dhabi Institute (AD131).

\ifCLASSOPTIONcaptionsoff
\newpage
\fi

{\small
\bibliographystyle{IEEEtran}
\bibliography{egbib}
}

%

\begin{IEEEbiography}[{\includegraphics[width=1in,height=1.25in,clip,keepaspectratio]{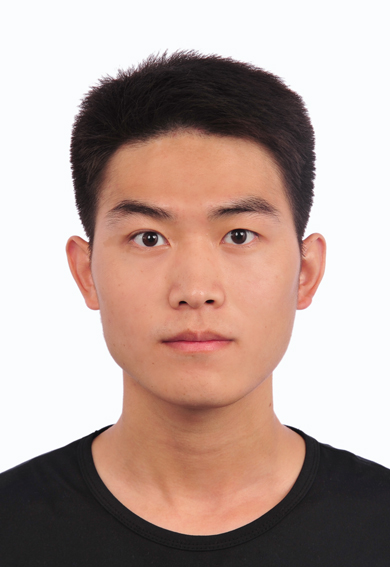}}]
{Xiang Li} received a B.S. degree in remote sensing science and technology from Wuhan University, Wuhan, China, in 2014. He received a Ph.D. in cartography and GIS from the Institute of Remote Sensing and Digital Earth, Chinese Academy of Sciences, Beijing, China, in 2019. 
He is currently a Postdoctoral Associate with the Department of Electrical and Computer Engineering, New York University Abu Dhabi, Abu Dhabi, United Arab Emirates. His research interests include deep learning, computer vision, and remote sensing image recognition.
\end{IEEEbiography}

\begin{IEEEbiography}[{\includegraphics[width=1in,height=1.25in]{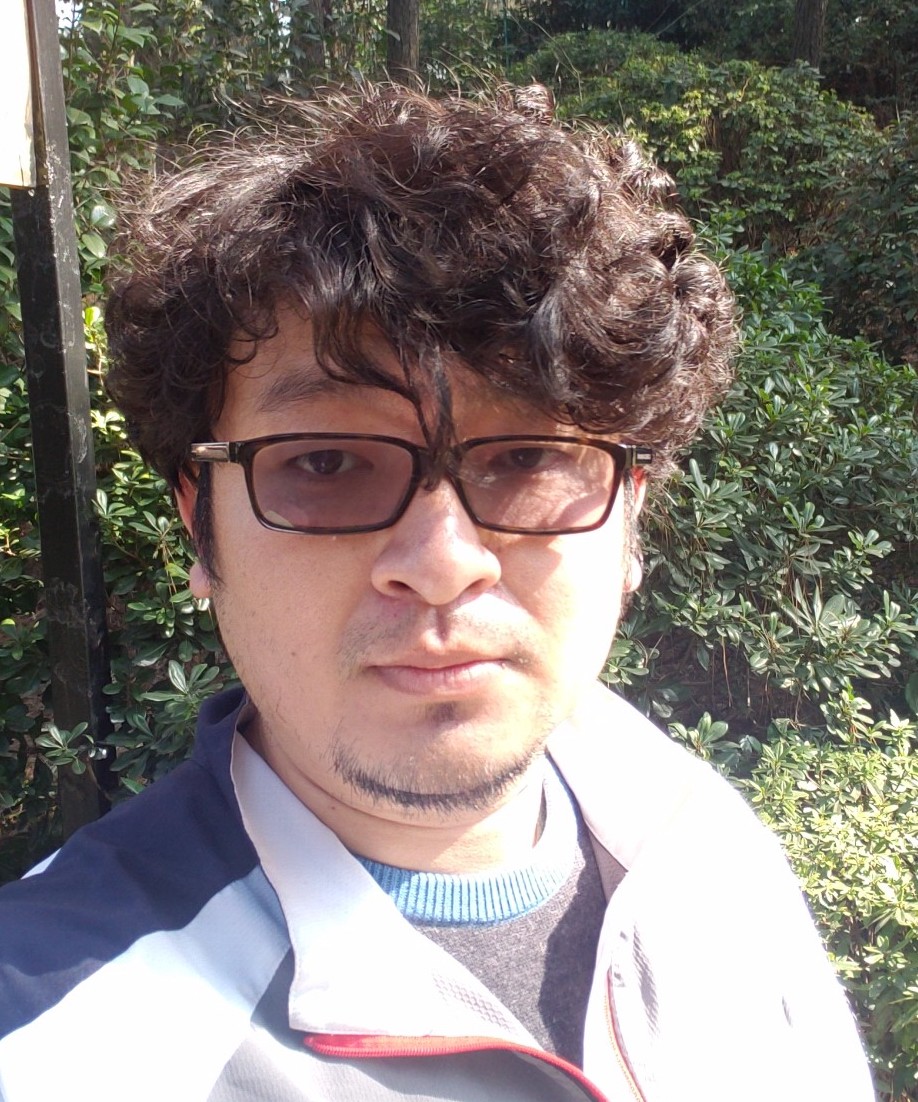}}]
{Lingjing Wang} received a B.S. degree from Moscow State University, Moscow, Russia, in 2011. He received a Ph.D. at the Courant Institute of Mathematical Science, New York University, USA, in 2019. 
He is currently a Postdoctoral Associate with the Department of Electrical and Computer Engineering, New York University Abu Dhabi, Abu Dhabi, United Arab Emirates. His research interests include deep learning and 3D visual computing.
\end{IEEEbiography}

\begin{IEEEbiography}[{\includegraphics[width=1in,height=1.25in]{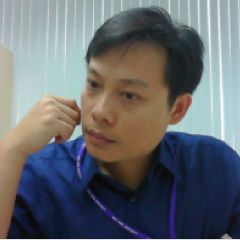}}]
{Yi Fang} received B.S. and M.S. degrees in biomedical engineering from Xi’an Jiaotong University, Xi’an, China, in 2003 and 2006, respectively, and a Ph.D. in mechanical engineering from Purdue University, West Lafayette, IN, USA, in 2011.
He is currently an Assistant Professor with the Department of Electrical and Computer Engineering, New York University Abu Dhabi, Abu Dhabi, United Arab Emirates. His research interests include three-dimensional computer vision and pattern recognition, large-scale visual computing, deep visual computing, deep cross-domain and cross-modality multimedia analysis, and computational structural biology.
\end{IEEEbiography}

\end{document}